\documentclass[wcp]{jmlr}

\usepackage{longtable}

\usepackage{booktabs}
\usepackage{fancyvrb} 
\usepackage{varwidth} 

\makeatletter
\let\Ginclude@graphics\@org@Ginclude@graphics 
\makeatother

\jmlrvolume{222}
\jmlryear{2023}
\jmlrworkshop{ACML 2023}

\title[Show Me How It's Done]{Show Me How It's Done: The Role of Explanations in Fine-Tuning Language Models}

\author{\Name{Mohamad Ballout} \Email{mohamad.ballout@uos.de} \\
     \Name{Ulf Krumnack} \Email{krumnack@uos.de} \\
     \Name{Gunther Heidemann} \Email{gheidema@uos.de} \\
     \Name{Kai-Uwe K\"uhnberger} \Email{kkuehnbe@uos.de} \\
      \addr Institute of Cognitive Science, University of Osnabrück, Osnabrück, Germany}

\editors{Berrin Yan{\i}ko\u{g}lu and Wray Buntine}

\begin{document}

\maketitle

\begin{abstract}
Our research demonstrates the significant benefits of using fine-tuning with explanations to enhance the performance of language models. Unlike prompting, which maintains the model's parameters, fine-tuning allows the model to learn and update its parameters during a training phase. In this study, we applied fine-tuning to various sized language models using data that contained explanations of the output rather than merely presenting the answers. We found that even smaller language models with as few as 60 million parameters benefited substantially from this approach. Interestingly, our results indicated that the detailed explanations were more beneficial to smaller models than larger ones, with the latter gaining nearly the same advantage from any form of explanation, irrespective of its length. Additionally, we demonstrate that the inclusion of explanations enables the models to solve tasks that they were not able to solve without explanations. Lastly, we argue that despite the challenging nature of adding explanations, samples that contain explanations not only reduce the volume of data required for training but also promote a more effective generalization by the model. In essence, our findings suggest that fine-tuning with explanations significantly bolsters the performance of large language models.

\end{abstract}

\section{Introduction}

Fine-tuning \citep{NEURIPS2022_0cde695b} and prompting \citep{arora2022ask, weichain} of large language models (LLMs) have recently become a vibrant field of study, driven by the successes of pre-trained language models. Current practices predominantly involve adapting readily accessible LLMs for both language-related \citep{ziegler2019fine, raffel2020exploring} and non-linguistic tasks \citep{lu2021pretrained, Ballout2023pretrain}. This trend necessitates the deployment of effective fine-tuning and prompting strategies to maximize the benefits of LLMs. 

Our study delves into the potential benefits of incorporating explanations into the output dataset to improve model performance. We discovered that the inclusion of explanations not only enhances performance on certain tasks but also empowers the model to learn and perform tasks it previously couldn't handle without explanations. We have developed a synthetic dataset, named Explained-ListOps-30, based on the original ListOps dataset from \citep{taylong,nangia2018listops}, which encompasses three distinct types of explanations: short, medium, and long. These explanations provide a step-by-step guide on problem-solving. Our findings indicate that all these explanation types significantly enhance performance, leading to nearly perfect scores (around 100\%) on tasks that previously achieved only about 65\% success rate without explanations.

Currently, there is a wide array of research approaches in this domain. For instance, some works are exploring methods such as in-context learning, where the model learns from a few examples provided in the prompt \citep{NEURIPS2020_1457c0d6, radford2019language, chen2022meta}. Conversely, other methods involve instruction-tuning, where the models receive specific guidelines to follow in fine-tuned tasks \citep{chung2022scaling,wang2022super}. These techniques are designed to enhance the model's proficiency in handling unseen tasks and complex reasoning problems.

Prompting techniques such as the `Chain-of-Thoughts' method \citep{weichain, zhang2023multimodal} are designed to guide the model to emulate human cognitive processes. In this approach, models are not just provided with examples and their corresponding solutions, but also with intermediate steps to illustrate the correct thought process leading to the answer. A more recent development expanding upon the chain-of-thought approach is the Tree-of-Thoughts method \citep{yao2023tree}. This approach empowers language models to engage in deliberative decision-making, allowing for the consideration of multiple reasoning paths and the ability to backtrack when a change in decision is warranted.

\begin{figure}[t]
\centering
\includegraphics[scale=0.35]{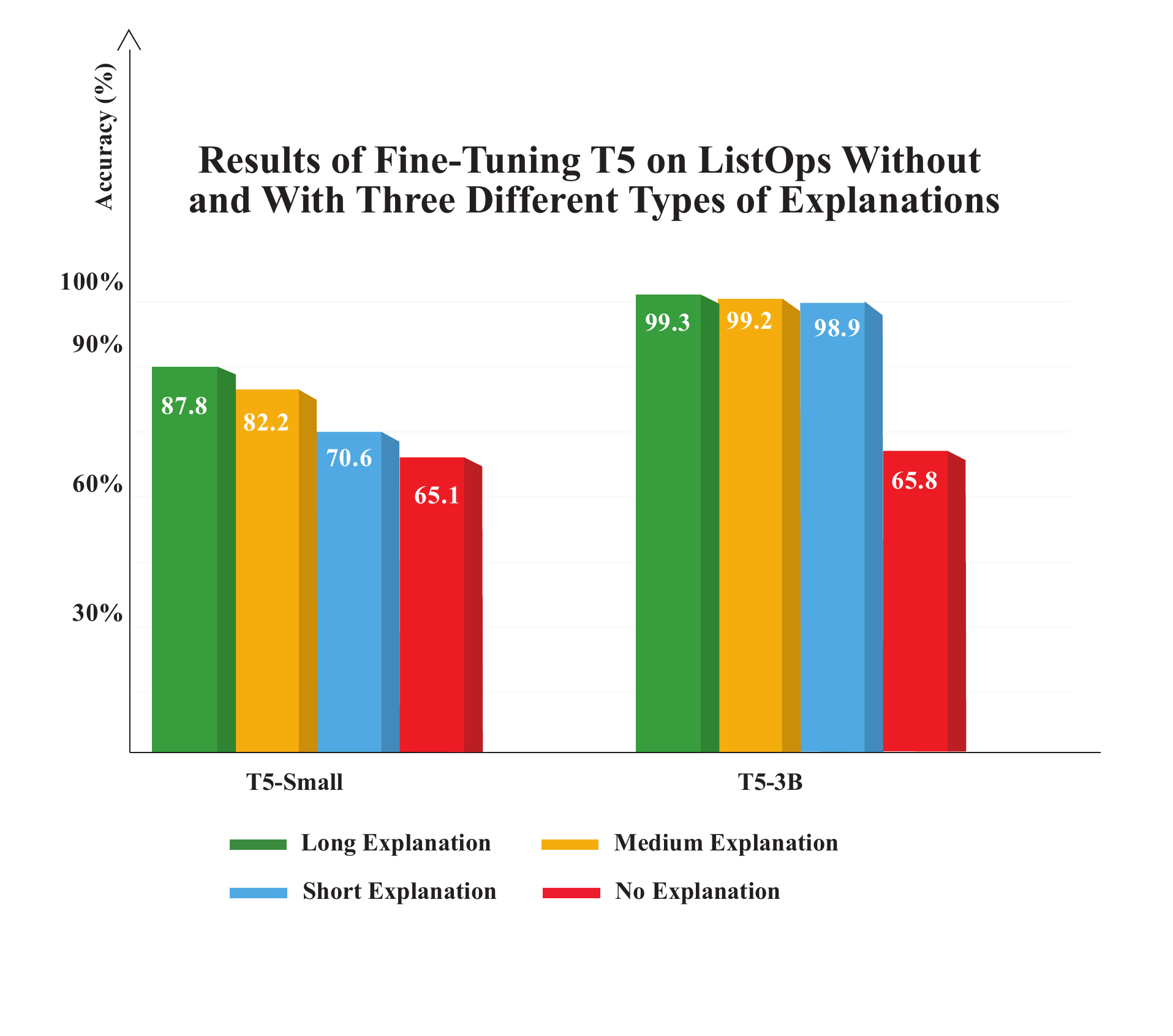}
\caption{Figure shows the results of the T5-small and T5-3B models on ListOps, comparing performances when fine-tuned without explanations versus with three the types of explanations.} 
\label{fig1}
\end{figure}

While existing methods yield impressive outcomes, \citet{wei2022emergent, weichain} have demonstrated that they primarily function effectively with significantly large models (100 billion parameters). In this study, we employ an approach similar to the chain-of-thoughts, but with fine-tuning on synthetic non-language tasks, namely ListOps. Rather than merely providing the answer in the output, we instruct the models step-by-step in three different manners to solve the problem. Upon publication, the code, model, and datasets used in our experiments will be made available as open source resources.\footnote{https://github.com/BalloutAI/Fine-tuning-with-Explanation.git}

The main contributions of this paper are:

\begin{itemize}
\item
We propose a fine-tuning approach that transforms a classification task into an answer generation task. Rather than simply providing the answer, we present intermediate steps to the models. Various types of steps, including short, medium, and long, are compared to determine the optimal explanation method. We discovered that all models, regardless of their size, benefit from different types of explanations. However, smaller models gain the most from lengthy and detailed explanations. 

\item

In contrast to established literature approaches, our method is effective with smaller language models. The results show that fine-tuning a model with 60 million parameters using our method scores of 87.8\% as shown in figure \ref{fig1}. This is significantly higher than a 3 billion parameter model without explanation, which scores around 65.75\% on the ListOps dataset. Our approach also empowers these smaller models to solve problems with a high accuracy of around 99\%, a dramatic improvement from previous results that were as low as 10\%, no better than a random guess.

\end{itemize}

\section{Literature Review}

This work is inspired by a prior study \citep{Ballout2023attention} where, through an examination of the attention weights of a fined-tuned pre-trained language model on ListOps, it was shown that the models were attempting to solve tasks in a manner akin to human problem-solving methods. Consequently, in this research, we explicitly present steps from human problem solving to the model and demonstrate that these steps support task resolution. To achieve this, we shifted the nature of tasks from classification problems, where the model is expected to predict a class, to generation tasks, wherein the model is required to generate the steps to solve the problem.

Many studies in the literature have generated datasets using natural languages, which were then used to fine-tune language models \citep{narang2020wt5, hase2020leakage, camburu2018snli}. While these works use explanations in their datasets, none have managed to improve model performance through the incorporation of these explanations. For instance, \citet{narang2020wt5} incorporated explanations into the T5 model; although their work increased the interpretability of the model, they were unable to surpass the original performance achieved without explanations. 

The work most similar to ours are \citet{hase2022can} and \citet{magister2022teaching}. \citet{hase2022can} used synthetic datasets with explanations to improve the performance of the language model. However, our research differs from theirs in that we examine different types of explanations and their effects on various model sizes.  In addition, we demonstrate that with the incorporation of explanations, our model is able to solve tasks that it was previously unable to solve with an accuracy beyond random chance. Furthermore, our work  bears similarities with the study by \citet{rajani2019explain}, where the authors evaluate common sense reasoning by fine-tuning models to explain answers. However, they were not able to show sustained improvements in model accuracy.
 
This study primarily focuses on fine-tuning models rather than in-context learning. Work also related to ours in existing literature includes research conducted by \citet{ling2017program} and \citet{cobbe2021training}. While \citeauthor{ling2017program} employs a similar strategy of generating explanations for mathematical problems instead of merely providing answers, they trained their model from scratch instead of using a pre-existing pre-trained language model. In contrast, \citeauthor{cobbe2021training}, although they fine-tune pre-trained language models on mathematical tasks, their method diverges from ours. They incorporate mathematical "verifiers", equations designed to assess the validity of the generated responses, into pre-existing explanations. This approach is distinct from ours wherein we provide a comprehensive step-by-step guide in the output to solve the problem.

\section{Datasets and Models}

\subsection{Datasets}
\label{sec:dataset}
To examine the effectiveness of different types of explanations, we developed a dataset which we call Explained-ListOps-30. The original ListOps \citet{taylong} is a hierarchical dataset designed to test a model's ability to handle such data structures. It involves a classification task involving ten classes, which represent the numbers from 0 to 9. These sequences incorporate mathematical operators such as Minimum, Maximum, Median, and Modular Sum. We kept the input the same as the original, but reduced the length to 50-100, down from the original 500-2000.  

The T5-base model scores approximately 70\% when the sequences are of length 50-100. However, we noticed that some shortcuts could be employed such as predicting 0 when the sequence begins with the Min operator, 9 with the Max operator, and 4 with the SUM and MED. Using this simple method, the model can attain an accuracy of around 35\%. To increase the complexity of the problem and reduce the impact of random guesses, we modified the classification task from 10 to 30 classes by expanding the range of integers from 0-9 to 0-29. If the model attempts the same trick as described above, the accuracy would drop to approximately 9\%. A short example of a sequence from this dataset could be:

$${[\text{MIN } \ 5 \ 20 \ [\text{MAX} \ 12 \ 26 \ [\text{MED} \ 1 \ 23 \ ] \ [\text{SUM} \ 15 \ 27 \ 1 ] \ ] \ ] }$$ 

This short sequence is comprised of 17 tokens, with each operator, number, and closed bracket being considered a token. In contrast, the actual dataset has a length of 50 to 100 tokens. In this scenario, the model should be able to determine that it must solve the sub-sequences in a specific order - first addressing the SUM operator, then the MED, MAX, and finally the MIN.

\begin{figure}[t]
\centering
\includegraphics[scale=0.7]{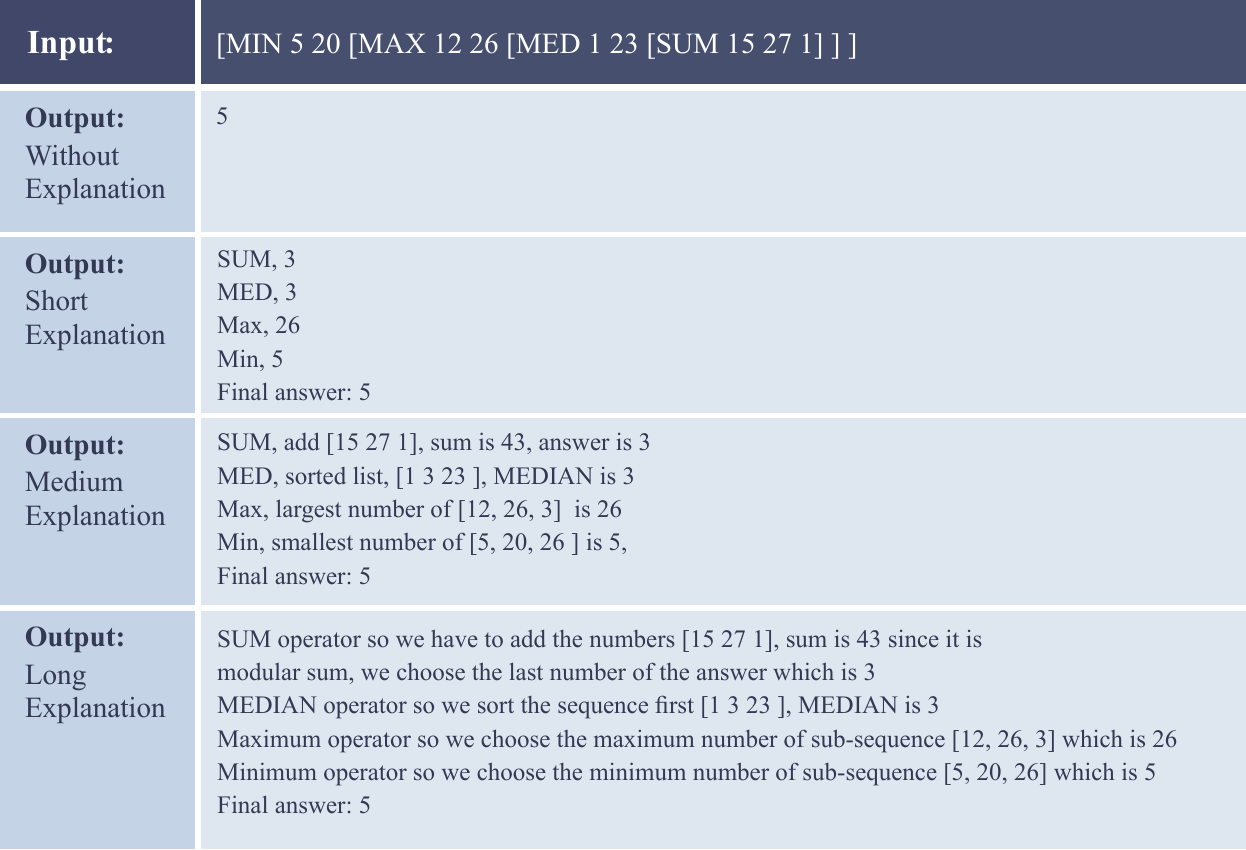}
\caption{Figure shows the three type of explanations produced for a short sample of the input} 
\label{fig2}
\end{figure}

In the Explained-Listops-30, we incorporated three types of step-by-step solutions, as depicted in figure \ref{fig2}. All these explanations provide intermediate solutions that solve the sub-sequences. A sub-sequence is the operator with the numbers inside the bracket. For instance, in the example above [MED 1 23] is a sub-sequence of the sequence.  The short explanation doesn't display the sub-sequence; it only shows the operator and the result for the sub-sequence. The medium-length explanation doesn't explain in plain language how to solve the problem, but it does present the sub-sequence to be solved and its operator. Finally, the long explanation offers a detailed, plain-language explanation of how to solve the sub-sequences until the final answer is achieved.

The aim of providing these explanations is to examine which type of explanation is most beneficial for LLMs and to address questions such as: are brief explanations as beneficial to the models as the long explanations, or is a complete instruction guide necessary?

\subsection{Models}

We choose to evaluate our explanations across a variety of models, ranging from very small language models to relatively larger ones. To ensure model consistency, we selected models from the T5 family \citep{raffel2020exploring}, including T5-small, T5-base, T5-large, and T5-3B. T5 is a text-to-text transformer, which does not differ much from the original encoder decoder model from \cite{vaswani2017attention}, designed to manage different types of tasks, such as translation, summarization, and more. A summary of each model's architecture and the number of parameters is presented in table \ref{tab:models}.

\begin{table}
\centering
\begin{tabular}{ccccc}
\hline
\textbf{Model} & \textbf{Parm.} & \textbf{Blocks} & \textbf{Emb.Size} & \textbf{Heads} \\
\hline
Small & {60M} & 6 & 512 & 8\\ 
Base & {220M} & 12 & 768 & 12\\ 
Large & {770M} & 24 & 1024 & 16\\ 
3B & {2.7B}  & 24 & 1024 & 32\\ 

\hline
\end{tabular}
\caption{Table shows the number of parameters and the different architectures of the models }
\label{tab:models}
\end{table}

\subsection{Fine-tuning Process}

In our experiments, we fine-tune the T5 models utilizing the Hugging Face implementation \citep{wolf2019huggingface}. We generate 98,000 training samples and 2,000 validation samples. The training process, adapted from \citet{nogueira2021investigating}, uses a learning rate of 3e-4. We fully fine-tune all parameters without freezing any layers. While the model is fine-tuned to generate intermediate steps, the reported accuracy is only based on the final answer. We fine-tune the model on four types of the ListOps dataset: the original ListOps without explanation, short Explained-ListOps-30, medium length Explained-ListOps-30, and long length Explained-ListOps-30. 

\begin{figure}[tbp]
\centering
\includegraphics[scale=0.85]{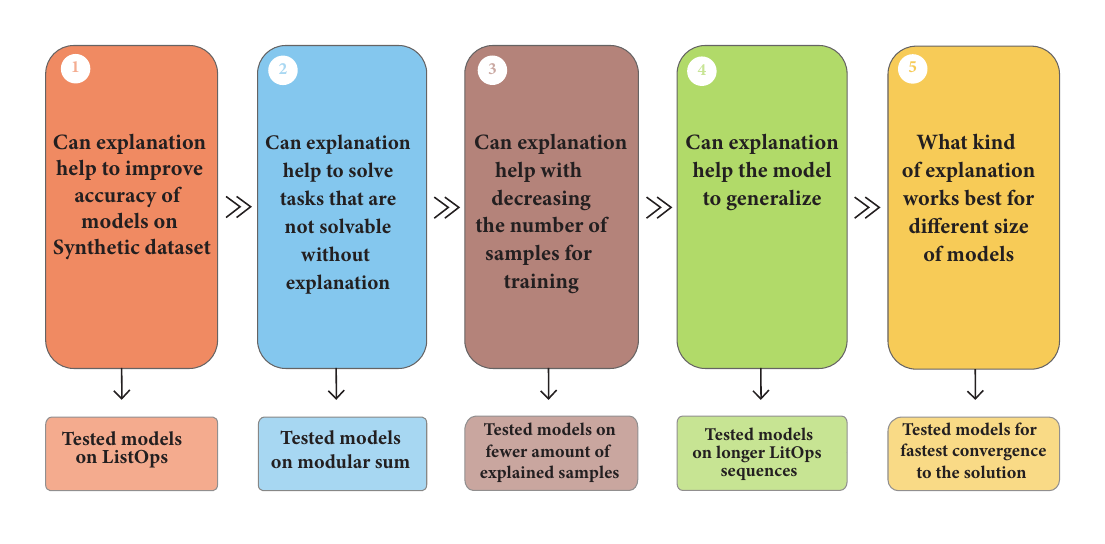}
\caption{Figure shows the experiments done in this paper to analyze the effect of incorporating explanations in the dataset} 
\label{fig3}
\end{figure}

In figure \ref{fig3} we show a summary of the analysis done in these experiments in order to study the effect of incorporating explanations in the dataset on the pre-trained language models and their performance.

\section{Results}
\subsection{Effect of Explanation on ListOps}
First, we fine-tune the four mentioned models on the four mentioned datasets. The results are shown in table~\ref{results}.

\begin{table}[htbp]
\centering
\begin{tabular}{lccccc}
\hline
\textbf{Model} & \textbf{No Exp.} & \textbf{Short} & \textbf{Medium} & \textbf{Long}\\
\hline
Small & 65.1\% & 70.6\% & 82.2\% & 87.8\% \\
Base & 65.3\% & 97.2\% & 99.6\% & 99.5\%\\
Large & 65.9\% & 98.7\% & 99.5\% & 99.5\%\\
3B & 65.8\% & 98.9\% & 99.2\% & 99.3\% \\

\hline
\end{tabular}
\caption{\label{results}
Results of fine-tuning the models on the explained and not explained ListOps.
}
\end{table}

Upon examining the data presented in the table, we can draw several conclusions. The results demonstrate that, without explanations, the performance of all the models, regardless of size, remains approximately the same. The models do not benefit from scaling up in size. For instance, the model with 60 million parameters performs as well as the model with 3 billion parameters. These findings are consistent with some literature results suggesting that upsizing the models is not always beneficial for certain types of tasks. Upon closer inspection of the results, we found that the models struggle to execute the modular sum operator, which we test separately later on.

The results indicate that all the models benefit from explanations in the output. However, it's apparent that the smaller model gains less benefit relative to the larger model. Additionally, the smaller model benefits more from longer explanations, scoring around 87.8\% compared to 70.6\% with shorter explanations. The larger models all benefit from any type of explanation, regardless of the level of detail, with only minor differences. For example, the accuracy of the T5-base with a short explanation is around 97.2\%, whereas with a medium explanation it is approximately 99.45\%. Similarly, for the large and 3B models, the difference is less than 1\%. From these findings, we observe that all LLMs improve with any type of step-by-step explanation, which are interesting results. All training is conducted over 30 epochs.

To ensure the generalizability of our results across various language models, we evaluated three additional models: BART \cite{lewis2020bart}, Pegasus \cite{zhang2020pegasus}, and Marian \cite{junczys2018marian}. The results presented in Table \ref{result-models} indicate that different types of models, which were pre-trained for diverse purposes and using varied methods, all demonstrate improvement when explanations are incorporated into their outputs. Although not all models achieve the level of improvement observed with T5, there is a significant increase in accuracy compared to outputs without explanations. We also observed that medium-length explanations yield the most consistent results across different models. Both the Pegasus and Marian models exhibit reduced performance with longer explanations, which we believe is due to their difficulty in handling an extended output of tokens.

\begin{table}[htbp]
\centering
\begin{tabular}{lccccc}
\hline
\textbf{Model} & \textbf{No Exp.} & \textbf{Short} & \textbf{Medium} & \textbf{Long}\\
\hline
T5-Base & 65.3\% & 97.2\% & 99.6\% & 99.5\%\\
BART & 63.1\% & 95.2\% & 97.0\% & 96.5\% \\
Pegasus & 66.1\% & 96.1\% & 95.6\% & 91.5\%\\
Marian & 61.1\% & 93.4\% & 95.7\% & 90.3\% \\

\hline
\end{tabular}
\caption{\label{result-models}
Results of fine-tuning the models on the explained and not explained ListOps.
}
\end{table}

\subsection{Effect of Explanation on Modular Sum}

In their study \citet{Ballout2023attention}, the authors determine that the modular sum operation (modulo 10) - where if, for instance, the sum is 43, the answer is the last digit of the number, which is 3 - presents a greater challenge than other operators. Thus, we chose to focus our tests on this operator to increase the task's complexity. We tested four models on sequences with lengths ranging from 50 to 100. An example of a short modular sum sequence might be:

$${[\text{SUM} \ 1 \ 20 \ [\text{SUM} \ 22 \ 6 \ [\text{SUM} \ 1 \ 23 \ [\text{SUM} \ 15 \ 27 \ 1 ] \ ] \ ] }$$ 

The answer without explanation would be 6, whereas a medium explanation of this sequence is: 

\begin{center}
\begin{BVerbatim}
SUM, add [15 27 1], sum is 43, answer is 3
SUM, add [1 23 3], sum is 27, answer is 7
SUM, add [22, 6, 7], sum is 35, answer is 5
SUM, add [1, 20, 5], sum is 26, answer is 6
Final answer: 6
\end{BVerbatim}
\end{center}

The other explanations (short and long)  with different length are similar to the ones shown in figure \ref{fig2}, but with having the modular sum operator only. The results are shown in table \ref{results2}. It's observed that none of the models without any explanation are able to solve the problem; the accuracy is on par with random guessing. We note that, in this task, the classification reverts back to 10 classes since the last digit is taken as the answer, reducing the number of classes to the digits from 0 to 9. With explanations, it can be seen that the models are able to solve the problem, with the exception of T5-small with short explanations. This reaffirms the conclusion drawn in the previous section: the T5-small model needs more detailed explanations to gain benefits as with short explanation the model does not learn better than the random chance whereas with medium and long explanation it scores 57.6\% and 75.7\% respectively. On the other hand, larger models can benefit from shorter, less detailed hints as well as long explanations. We note that these experiments are conducted over 100 epochs for the models showing that the task is more challenging to solve.  

\begin{table}[htbp]
\centering
\begin{tabular}{ccccc}
\hline
\textbf{Model} & \textbf{No Exp.} & \textbf{Short} & \textbf{Medium} & \textbf{Long}\\
\hline
Small & 11.3\% & 11.3\% & 57.6\% & 75.7\% \\
Base & 11.2\% & 98.1\% & 99.6\% & 99.2\%\\
Large & 10.9\% & 99.4\% & 99.1\% & 98.9\%\\
3B & 11.0\% & 98.3\% & 99.1\% & 99.5\% \\

\hline
\end{tabular}
\caption{\label{results2}
Results of fine-tuning the models on the explained and not explained modular sum.
}
\end{table}

\subsection{Can the Model Learn With Less Number of Samples?}

In this section, we investigate whether the inclusion of explanations in the data reduces the number of necessary samples. For this purpose, we train the model using just 2,000 explained samples. The results shown in table \ref{tab:2000} reveal that the model was able to learn more from 2,000 explained examples than it could from 98,000 samples without explanations. With a data size of 2,000 examples, the T5-Large model trained on explained samples achieved an accuracy of 66.1\% , while with 2,000 unexplained data samples, the model's accuracy is 44.9\%. We note that the accuracy achieved through random guessing is approximately 3.3\%, while the simple shortcut described in Section \ref{sec:dataset} yields around 9\% accuracy. Evidently, the performance of the medium and long explanation surpasses both of these methods by a significant margin.

\begin{table}[htbp]
\centering
\begin{tabular}{lccc}
\hline
\textbf{Model} & \textbf{No explanation} & \textbf{Medium Expl.} & \textbf{Long Expl.} \\
\hline
T5-Large & 44.9\% & 66.1\% & 65.9\%\\ 
\hline
\end{tabular}
\caption{Results of the T5-large model when trained on 2000 samples of with medium length, long and no explanation}
\label{tab:2000}
\end{table}

\begin{figure}[h]
\centering
\includegraphics[scale=0.85]{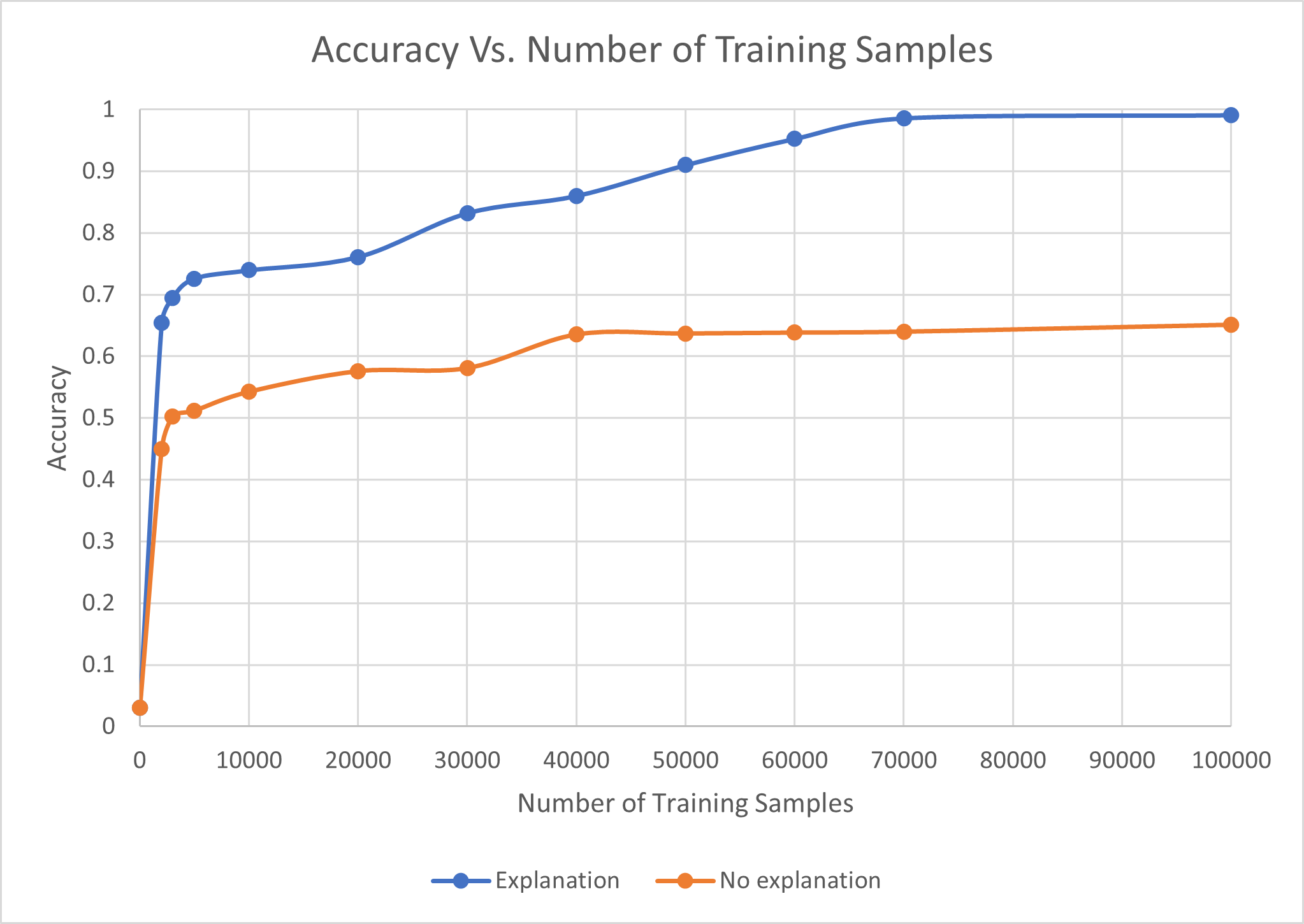}
\caption{The performance achieved by T5-large at end of training for training sets of different sizes, with medium sized explanations and without explanations.} 
\label{fig6}
\end{figure}

Finally, as illustrated in figure \ref{fig6}, we plot the relationship between accuracy and the number of samples used. As demonstrated in table \ref{tab:2000}, when trained with 2000 samples and provided an explanation, the model achieves an accuracy rate of 66.1\%. As more samples are introduced to both explained and unexplained datasets, there is a corresponding increase in accuracy, culminating in the final accuracy achieved at around 100,000 samples. The results depicted in figure \ref{fig6} align with expectations. Initially, the model derives substantial learning from a mere 2000 samples. Following this, the accuracy demonstrates a steady ascent as additional explained and unexplained samples are incorporated, eventually plateauing at a threshold. It's worth noting that figure \ref{fig6} only displays the medium explanation due to the computational cost of the long explanation. This decision was made especially after table \ref{tab:2000} showed that similar results to the medium explanation are obtained when trained on 2000 samples.

\subsection{Can Explanation Help the Model to Generalize?}

In this section, we examine the model's ability to generalize to longer sequences, leaving the generalization to larger numbers to be discussed in the limitations section. The results show that T5-large model, trained with medium-length explanations on sequences of 50-100 and tested on sequences of length 100-200, achieves an accuracy of 91.2\%. In contrast, the generalization accuracy without explanation is around 63.5\%, which indicates that explanations enable the model to better generalize to longer sequences. We chose not to test the long explanation on sequences ranging from 100-200 in this section because the resulting answer might contain a large number of tokens, which the T5 model cannot accommodate.

\begin{table}[htbp]
\centering
\begin{tabular}{lcc}
\hline
\textbf{Model} & \textbf{No explanation} & \textbf{Medium Expl.} \\
\hline
T5-large & 63.5\% & 91.2\%\\ 
\hline
\end{tabular}
\caption{ Results of generalization test trained on sequences of length 50-100 and tested on sequences from 100-200}
\label{tab:gener}
\end{table}

\subsection{What Kind of Explanation Works Best for Different Size of Models?}

The effect of explanation on performance is clear when compared with models fine-tuned on unexplained data, and the difference between the types of explanations is evident with the T5-small model - the longer the explanation, the better the performance. However, the distinction between explanations for the other models, namely T5-Base, T5-Large, and T5-3B, remains unclear. Therefore, this section aims to analyze how fast each model converges and the amount of training needed to reach the solution with each explanation. This is a critical issue due to the considerable computational demands of large language models owing to their size. Consequently, providing insights into the most rapidly converging methods is vital for reducing computational requirements and promoting a training process with a smaller carbon footprint.

\begin{figure}[htbp]
    \centering
    \includegraphics[scale=0.65]{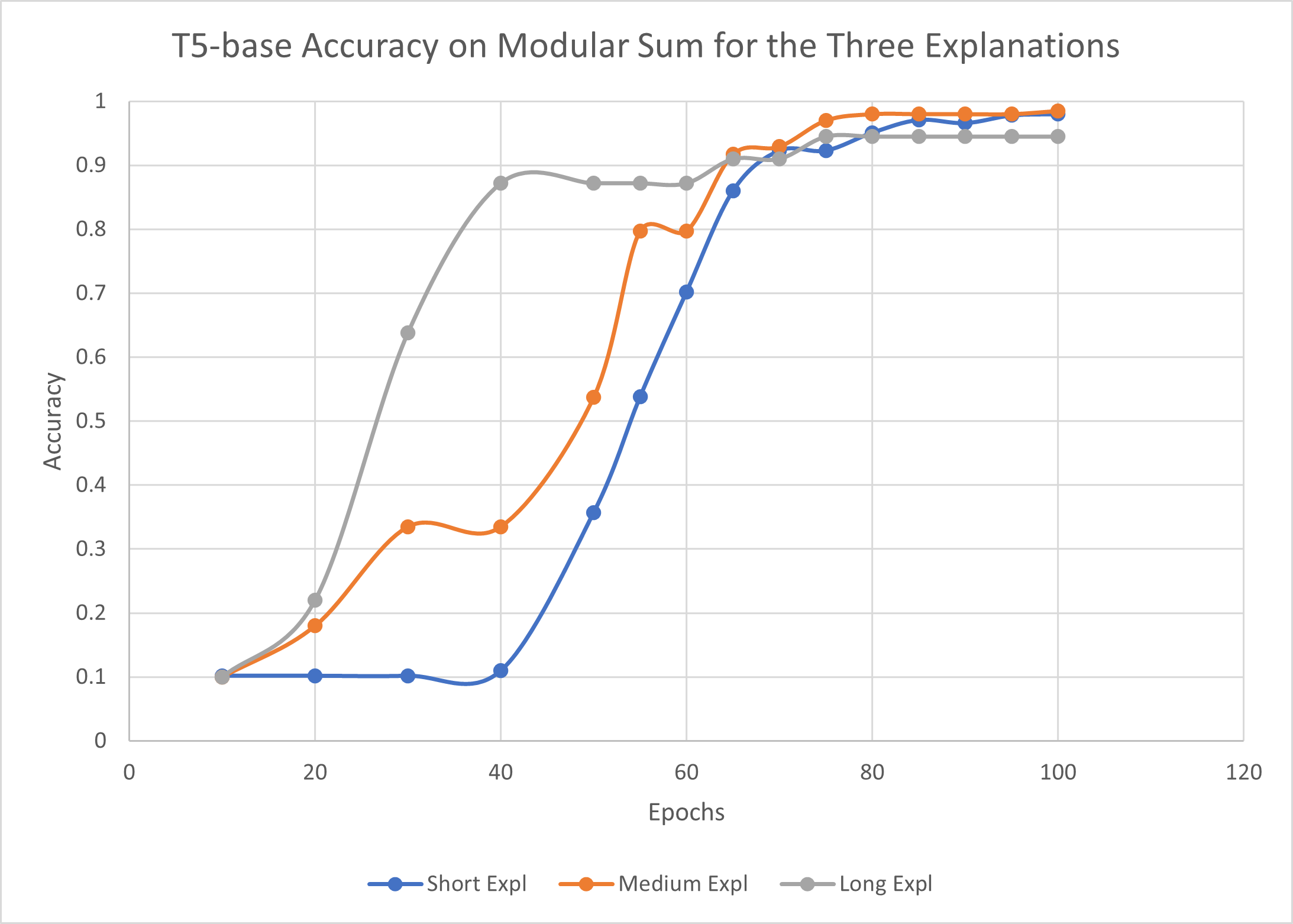}
    
    \vspace{1cm} 
    
    \includegraphics[scale=0.65]{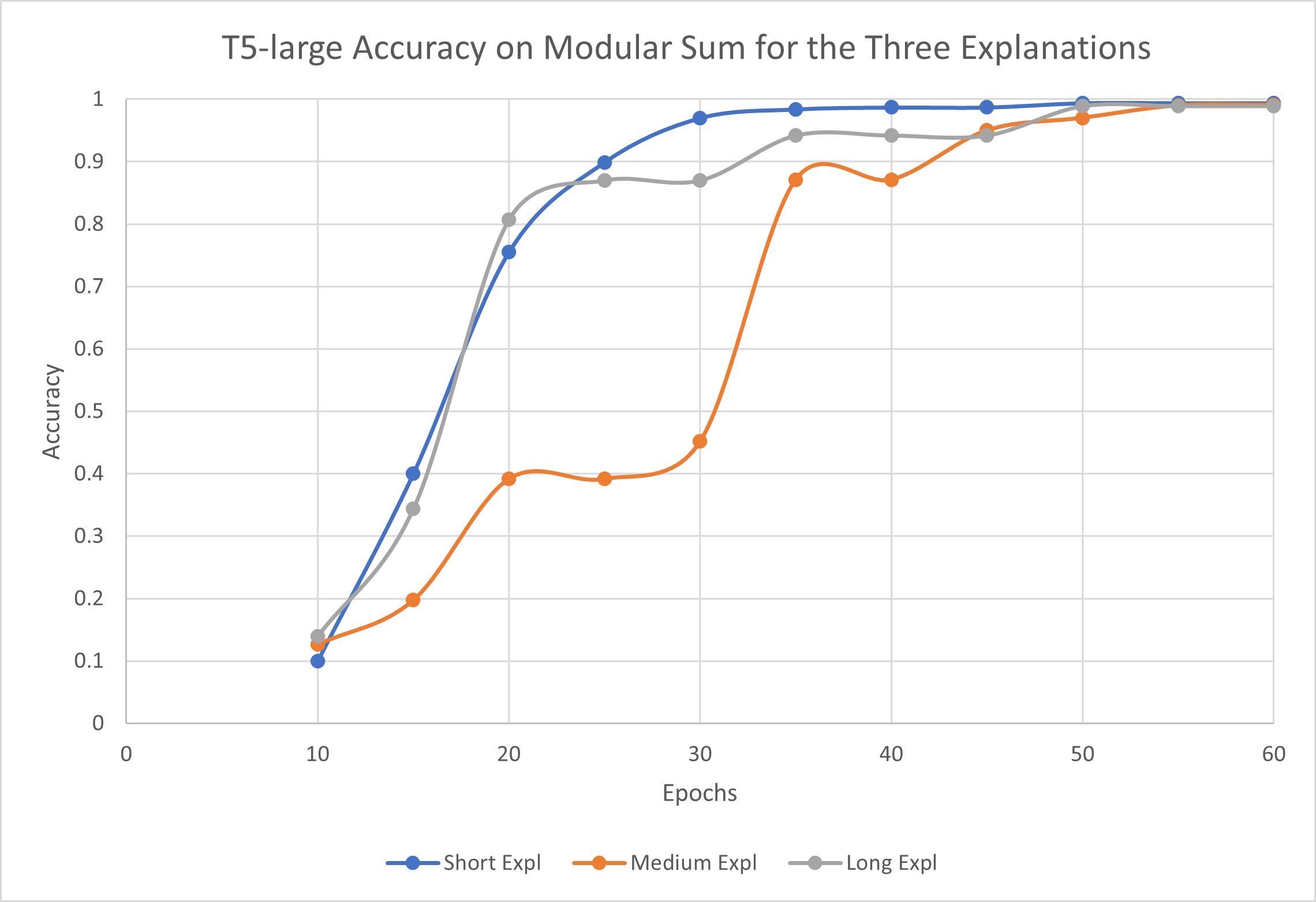}
    \caption{The convergence graph of the T5-base (top) and T5-large (bottom) models with three different explanations}
    \label{fig:both} 
\end{figure}

Since the benefits of each type of explanation on T5-small is obvious from the accuracy, we only show the accuracy plot of T5-base and T5-large while the same observations of T5-large are for the T5-3B. We can see from the graphs shown in figure \ref{fig:both} that the effect of the length of the explanation on T5-base is higher, while they all achieve similar accuracies, it is obvious from figure, where we show accuracy  on the ListOps dataset versus epochs numbers with different explanations that the longer the explanation is the faster it converges to the solution. On the other hand, this effect is less obvious with the larger model where there is no advantage of medium and long explanations over the short explanation. We can conclude a similar conclusion as from above where we stated that the larger the model is, the less need for detailed explanation in the dataset is needed, whereas the smaller versions of the models need more detailed explanations.

\section{Limitations}
In this section, we highlight a limitation of the proposed explanation method. We investigate whether this method assists in generalizing to larger numbers. As such, the generalization test involves training the model on sequences of numbers from 0 to 9 and testing it on numbers from 0 to 29. The results, displayed in the table \ref{tab:limitations}, indicate that the model is unable to generalize to sequences of larger numbers, regardless of the presence of explanations.

\begin{table}[htbp]
\centering
\begin{tabular}{lcc}
\hline
\textbf{Model} & \textbf{No explanation} & \textbf{Medium Expl.} \\
\hline
T5-Base & 31.8\% & 33.05\%\\ 
T5-3B & 30.35\% & 31.6\%\\ 
\hline
\end{tabular}
\caption{ Results of generalization test trained on numbers from 0-9 and tested on numbers from 0 -29.}
\label{tab:limitations}
\end{table}

\section{Conclusion and Future Work}

In this work, we fine-tuned T5 models of various sizes on differently explained versions of the ListOps dataset, as well as on the unexplained dataset. Our findings indicate that all forms of explanations significantly boost the models' accuracy. Moreover, these explanations enables the model to solve problems that were previously as challenging as making a random guess. We also discovered that while smaller models require more than just short explanations to improve, larger models, on the other hand, benefit from any explanation, irrespective of its length or details. In addition, we showed that incorporating explanations in the dataset helps the model to generalize to longer sequences, whereas it was not able to do so with a dataset that does not include explanations. Finally, we showed that with small models, longer explanations help them converge faster to higher accuracy as opposed to shorter explanations, which take more epochs to converge.

We believe that this research could offer valuable guidance on how to generate effective instructions and explanations, tailored to the specifics of the pre-trained language models being used. An extension of this work could involve testing the results presented in this paper on non-synthetic datasets.

\section*{Acknowledgements}

This work was funded by the Deutsche Forschungsgemeinschaft (DFG, German Research Foundation). The cluster used to train the models was also funded by the German Research Foundation (DFG) - 456666331.

\bibliography{acml23}
\end{document}